\documentclass[review]{elsarticle}

\usepackage{lineno,hyperref}
\modulolinenumbers[5]

\usepackage{graphicx}
\usepackage{amssymb}
\usepackage{amsmath}
\usepackage{tabularx}
\usepackage{lscape}
\usepackage{ltablex}
\usepackage{url}
\usepackage{color}
\usepackage{type1cm}

\usepackage[whole]{bxcjkjatype} 

\journal{Neural Networks}

\begin{document}

\begin{frontmatter}

\title{
The whole brain architecture approach: Accelerating the development of artificial general intelligence by referring to the brain}

\tnotetext[mytitlenote]{}

\author[WBAI,UT,RIKEN]{Hiroshi Yamakawa}

\cortext[cor]{Corresponding author}
\address[WBAI]{The Whole Brain Architecture Initiative, Nishikoiwa 2-19-21, Edogawa-ku, Tokyo, 133-0057, Japan}%
\address[UT]{The University of Tokyo, 7-3-1 Hongo, Bunkyo-ku, Tokyo 113-0033, Japan}
\address[RIKEN]{RIKEN, 6-2-3, Furuedai, Suita, Osaka 565-0874, Japan}%

\begin{abstract}

The vastness of the design space created by the combination of a large number of computational mechanisms, including machine learning, is an obstacle to creating an artificial general intelligence (AGI). Brain-inspired AGI development, in other words, cutting down the design space to look more like a biological brain, which is an existing model of a general intelligence, is a promising plan for solving this problem. However, it is difficult for an individual to design a software program that corresponds to the entire brain because the neuroscientific data required to understand the architecture of the brain are extensive and complicated. The whole-brain architecture approach divides the brain-inspired AGI development process into the task of designing the brain reference architecture (BRA)---the flow of information and the diagram of corresponding components---and the task of developing each component using the BRA. This is called BRA-driven development. Another difficulty lies in the extraction of the operating principles necessary for reproducing the cognitive-behavioral function of the brain from neuroscience data. Therefore, this study proposes the Structure-constrained Interface Decomposition (SCID) method, which is a hypothesis-building method for creating a hypothetical component diagram consistent with neuroscientific findings. The application of this approach has begun for building various regions of the brain. Moving forward, we will examine methods of evaluating the biological plausibility of brain-inspired software. This evaluation will also be used to prioritize different computational mechanisms, which should be merged, associated with the same regions of the brain.

\end{abstract}


\end{frontmatter}


\section {Introduction}
\label{sec:Introduction}


The development of artificial general intelligence (AGI), which is the goal of advanced artificial intelligence (AI) research in recent years, is to develop and demonstrate the extensive general intelligence possessed by humans within a computational system \citep{Adams2012-xs, Goertzel2014-oy}. An exceptionally essential ability of AGI would perhaps involve solving various problems, including those on unknown issues, by flexibly combining knowledge gained from experience 
 However, the method to develop AGI remains unclear. Many AI researchers believe that the development of deep learning \citep{LeCun2015-bf} was a launch pad towards this goal. According to them, this goal can be realized by combining various computational mechanisms, including machine learning, a method that allows a machine to learn knowledge from experience. Certainly, there have been attempts to create a unified theory and principle of intelligence \citep{Friston2010-tr, Domingos2015-su, Hafner2020-uo}. However, there is no single theory on which the entire sphere of intelligence can be built. Thus far, the development of AGI has made progress by repetitively tuning various limited issues. Nevertheless, such an approach would make it difficult to design an AGI with a flexible problem-solving ability that would enable it to solve unknown issues. Building an AGI that has the full extent of human abilities would require an extremely large design space as a result of combining computational mechanisms. This design space could also be explored mechanically \citep{Clune2019-fi}, but for the time being, it is difficult to secure the required computational complexity.


Brain intelligence is associated with a certain degree of versatility. If we can narrow down the design space by referring to the architecture of cognitive and behavioral functions in the brain \citep{Petersen2015-cn}, we can accelerate the development of an AGI that is comparable to human intelligence. Simply put, even if the scope of AGI realized by machines (machine kingdom) does not need to be bound by biological constraints \citep{Hernandez-Orallo2017-hr}, the development of a brain-like architecture could be an agreed-upon milestone in AGI development \citep{Hassabis2017-by, Goertzel2010-hy}.


Based on this technological background, we at the Whole Brain Architecture Initiative (WBAI) have been advocating for a development approach called the whole brain architecture  (WBA) approach since 2015. We define the basic idea of this AGI development approach as creating "a human-like artificial general intelligence by learning from the architecture of the entire brain" \citep{ Yamakawa2016-wf, Arakawa2016-ly} 


The premise of this approach is that the functions of the brain can be realized by combining several machine learning devices with well-defined functions. By imitating the structure of the brain and combining machine learning devices that imitate each brain function, it is possible to build an AGI with abilities that are similar to or beyond those of humans. This approach, therefore, uses a WBA-centered hypothesis. Based on this hypothesis, a machine learning module is developed for each part of the brain and the software is constructed by integrating these modules.

\begin{figure}[tb]
    \begin{center}
        \includegraphics[width=0.9\linewidth]{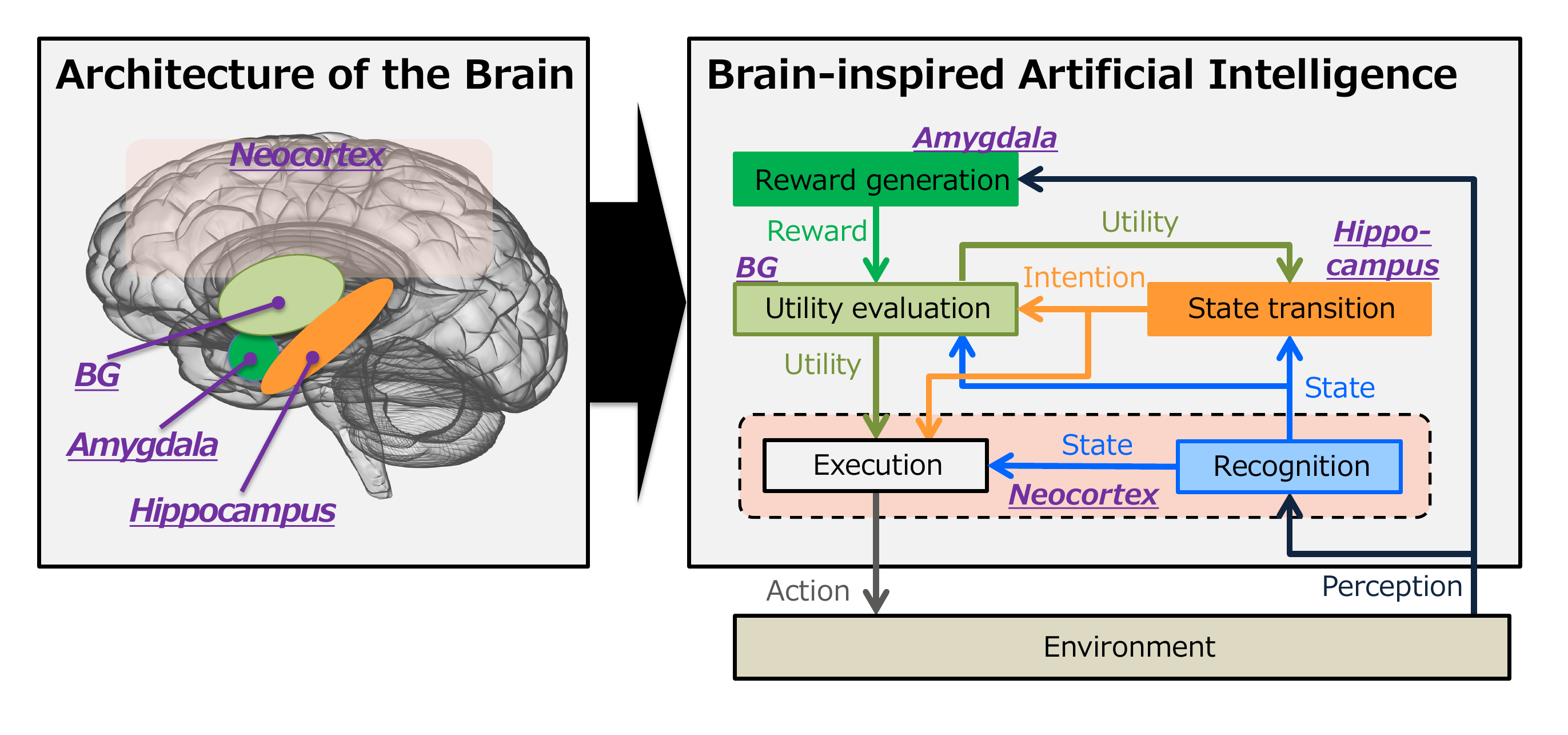}
    \end{center}
    \caption{Basic scheme of a whole-brain architecture approach.
   This is a revised version of a simplified diagram of the basic idea of the WBA approach, which has been gradually formed since around 2014. The left side of the figure highlights major examples of large organs in the brain, including the neocortex, basal ganglia (BG), hippocampus, and amygdala. An additional brain architecture is formed by the connections between these organs (not shown). On the right side of the figure, computational modules, including those utilizing machine learning, are placed and connected with reference to the architecture of the brain.  Building an artificial intelligence software system that can operate while interacting with the environment through the body is based on this scheme.
    }
    \label{fig:1}
\end{figure}

Based on these assumptions, the WBA approach aims to build a brain-inspired AGI based on the following basic ideas.
As shown in Fig.\ref{fig:1}, each brain organ is implemented as a calculation module, including those that utilize machine learning, and is integrated based on the architecture of the brain. The brain organs shown in the figure are fairly coarse, but in reality they are associated with the brain in units of  finer-grained computational modules (see \ref{subsec:Mesoscopic-level}).

It is not realistic to construct a brain-inspired AGI software by directly referring to neuroscientific findings in academic papers and data. This is because the functions of the brain are diverse, and the neuroscientific findings about these functions are vast. Furthermore, the number of people who thoroughly understand neuroscience and can develop software are limited, as it involves intensive training.

\begin{figure}[tb]
    \begin{center}
        \includegraphics[width=0.6\linewidth]{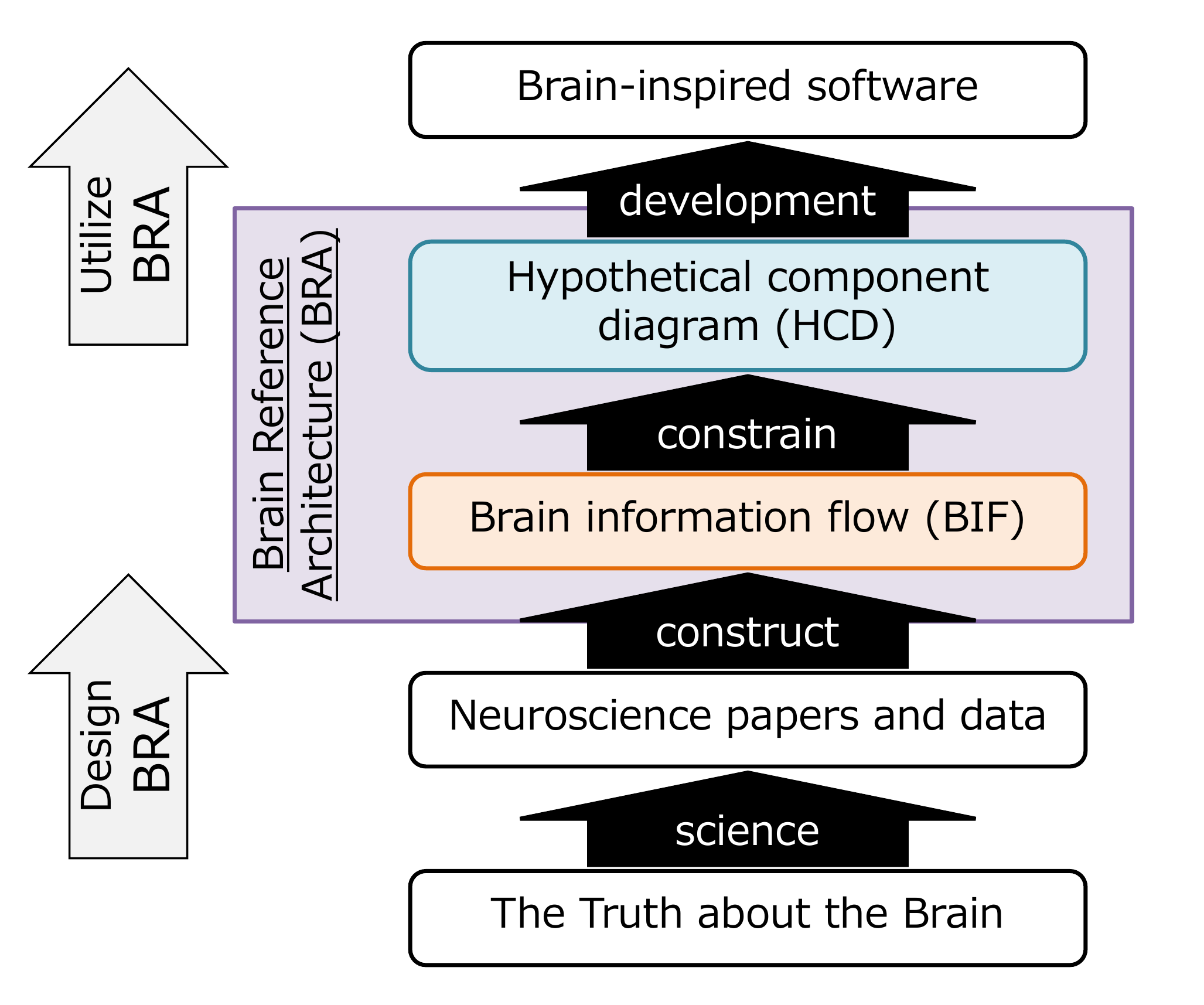}
    \end{center}
    \caption{BRA-driven development, which consists of the work of developing brain-inspired software using the BRA and the work of designing the BRA based on neuroscientific knowledge (papers and data). The BRA consists of the brain information flow (BIF) and the hypothetical component diagram (HCD).
    }
    \label{fig:2}
\end{figure}

To overcome this problem, WBAI standardizes information corresponding to the requirements for developing brain-inspired software as brain reference architecture (BRA) data. The methods of BRA design and utilization are shown in Fig. \ref{fig:2} \citep{Sasaki2020-zp}. We refer to this as BRA-driven development.

The BRA is a mesoscopic-level reference architecture (discussed later). It contains a description of the brain information flow (BIF) within the anatomical structures that is consistent with the hypothetical component diagram (HCD) related to computational functions. The BIF is a directed graph made up of partial circuits and connections representing the anatomical structure of neural circuits. The HCD is a representation of a directed graph that details the dependency relationship of components and can correspond to any subgraph in BIF (see Subsection \ref{subsec:HCD}  for details).


Therefore, even a developer with limited knowledge of neuroscience can implement the software provided that the BRA is given, and the HCD contained in the BRA is regarded as a requirement.


The development of brain-inspired software requires HCDs that cover all the regions of the brain. However, creating such HCDs is not an easy feat. Even with the recent, rapid developments in neuroscience, the operating principles have been described for only a few regions in the brain. To overcome this situation, we have developed the Structure-constrained Interface Decomposition (SCID) method. This method allows the creation of an HCD while compensating for the lack of information on operating principles. One advantage of SCID is that it is able to design HCDs consistent with the anatomy of a fairly wide area of the brain. Therefore, the SCID method is a powerful technique to help us create a BRA that covers the whole brain.
An AGI can make effective inferences in a situation that it has not directly experienced. Therefore, it must have the ability to efficiently acquire knowledge common to various tasks from finite experience, while focusing on various parts of the world, and to flexibly combine them. However, software developed with reference to different HCDs can only be implemented for each task that the respective HCD assumes; therefore, acquiring and using knowledge for tasks apart from those tasks tend to be difficult. This situation needs to be overcome in AGI development. In this regard, the WBA approach will use merge development, which integrates multiple different computational mechanisms related to the same brain regions with a certain granularity. In the near future, BRA-based developments will proceed alongside a parallel or iterative development that will merge the deliverables of BRA-based developments.

As BRA-driven developments progress, their deliverables tend to move away from the reality of the brain, which is a problem because the WBA approach aims to explore AGI within a design space close to that of the brain. To circumvent this problem, it is necessary to continue evaluating how well the implemented brain-inspired software reproduces the truth of the brain (the parts related to the level of cognitive behavior) perceived by neuroscience. The evaluation of such biological plausibility is carried out from two viewpoints. 
The first evaluation is adequacy for BRA to assess whether it is consistent with existing neuroscience findings.  The other is the evaluation of fidelity or whether the software is built according to the BRA \citep{Yamakawa2020-wa, Yamakawa2020-qz}.


Section 2 delves into the BRA and discusses the description level at which the brain-inspired AGI should learn about the brain; the BIF format, which is an element for describing the BRA; and the HCD. Furthermore, we will also provide an overview of BRA-driven development. Section 3 discusses the collection of neuroscientific findings related to BRA design, SCID, and the evaluation of BRA validity. Section 4 discusses the stub-driven development and merge development using the BRA and methods of evaluating the fidelity of software from the perspective of the BRA in the future. Finally, these discussions are summarized.

\subsection { Related AGI development approaches }
\label{subsec:BIF}


At the end of 2020, there exists at least 72 projects involved in AGI development; looking at the changes since 2017, the overall number has not changed much. However, 15 new projects, mostly from the private sector, have been added during this period\citep{Fitzgerald2020-ma, Baum2017-ls}.

In many AI projects, the researcher designs, implements, and evaluates a combination of computational mechanisms that enable the software to perform cognitive-behavioral functions. Projects following the traditional flow that starts with symbolic AI have a large design burden. Examples of these problems are ACT-R  \citep{Anderson2009-rb}, SOAR \citep{Laird1987-fs}, ICARUS \citep{Choi2018-tu}, LIDA  \citep{Franklin2014-ki}, NARS  \citep{Wang2018-bk}, Sigma \citep{Rosenbloom2016-lz}, CLARION    \citep{Sun2016-ig}, and CogPrime \citep{Goertzel2012-qp}. Hierarchical Temporal Memory \citep{Hawkins2004-nv, Krestinskaya2018-ih} is an advanced AGI development project that mainly started with mimicking neocortical functions in the brain using an artificial neural network and the brain's learning mechanism \citep{Hawkins2004-nv, Krestinskaya2018-ih}. Part of that flow was taken over by the Vicarious project \citep{George2018-cf}. Many projects started around 2015, when steady strides in the development of deep learning began. For example, in addition to OpenAI \citep{Brown2020-mb}, which produced remarkable language model results in recent years, there is GoodAI, NNAISENSE, and WBA. Founded in 2010, DeepMind has been developing AGI centered on deep learning and reinforcement learning while referencing the brain with a relatively high level of abstraction \citep{Mnih2013-zg, Silver2018-ua}. The Nego project \citep{Eliasmith2013-ib, DeWolf2020-np} is attempting to mimic the entire cognitive-behavioral function of the brain at the level of spiking neurons; however, the project is not considered to be explicitly challenging AGI \citep{Fitzgerald2020-ma}.

Related research areas to explore cognitive systems include biologically inspired cognitive architecture  \citep{Samsonovich2016-tn, Goertzel2010-hy} and cognitive computational neuroscience \citep{ Kriegeskorte2018-xw}, an interdisciplinary field of cognitive science and computational neuroscience, among others. However, these fields have not made progress in accumulating design data in a standardized way like the BRA.


Given that the brain is a type of natural object, it is rational to design AI by imitating physical processes that are similar to physical simulation and to evaluate it by how well the physical stimulation has been reproduced. In practice, studies on whole-brain simulation have been carried out with this perspective \citep{Bostrom2008-sk, Markram2006-ur}. If the physical processes that support cognitive-behavioral function can be fully elucidated and a basic equation can be developed, it is highly possible that human-like intelligence can be reproduced through such an approach. However, neuroscience, as it is right now, has not yet reached this level of detail, and it is hard to tell when it will achieve this. This is because it is still difficult to observe the neural activity of the whole brain with sufficient spatiotemporal resolution. Even if the activity can be observed, it remains difficult to connect complex processes in the living body to functional operating principles.

\section {Brain Reference Architecture}
\label{sec:BRA}

\begin{figure}[tb]
    \begin{center}
        \includegraphics[width=0.98\linewidth]{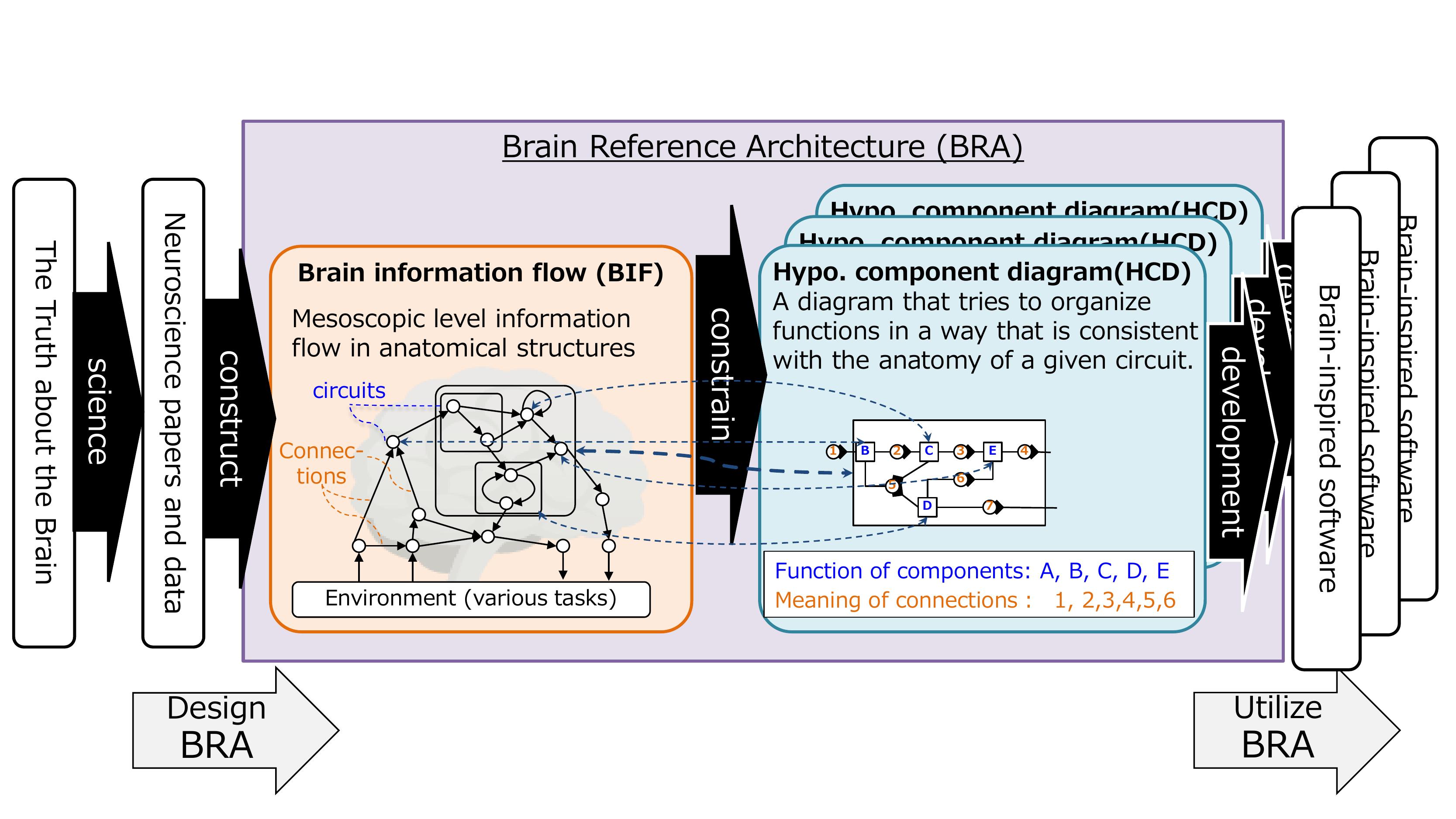}
    \end{center}
    \caption{Brain reference architecture.
    The BRA consists of the BIF and the HCD.  The BIF is the mesoscopic level information flow in anatomical structures.  The HCD is a diagram that organizes functions in a way that is consistent with the anatomy of a given circuit. There can be multiple HCDs for a single circuit for the BIF. Each software development project is essentially based on a specific HCD.
    }
    \label{fig:3}
\end{figure}

The BRA is a reference model that aggregates neuroscientific findings (on anatomical structures or physiological phenomena) on the mesoscopic-level architecture of the brain and gives a hypothesis of computational function consistent with those findings. Specifically, as shown in Fig. \ref{fig:3}, the data is a combination of BIF (see Subsection \ref{subsec:BIF}) and HCD (see Subsection \ref{subsec:HCD}) \citep{Sasaki2020-zp}.

The current WBA approach involves BRA-driven development. This development consists of a BRA design based on neuroscientific findings and BRA utilization that implements and evaluates its processes by referring to the HCD in the BRA.


After describing the BRA at the mesoscopic level, we will discuss the BIF as a component of the BRA and HCD, which are designed to be consistent with the BIF. We will also discuss BRA-driven developments

\subsection { Referenced mesoscopic level }
\label{subsec:Mesoscopic-level}


What part of neuroscience should the WBA approach refer to?
Generally, the following can be said about the granularity that should be referred to when trying to reproduce the cognitive-behavioral function of the brain. Coarse-grained references to large regions of the brain (e.g., neocortex, hippocampus, basal ganglia, and cerebellum) are too weak to recreate cognitive-behavioral functions of the brain. However, conversely, trying to reproduce the details inside the neuron (e.g., metabolism) complicates the construction of the design and increases the computational resources required to run a model that reflects this information, resulting in great disadvantages. Therefore, the reference granularity in the WBA approach should be assumed to be at the mesoscopic level.

\subsubsection{Uniform circuit: Finest reference granularity for brain architecture}

\begin{figure}[tb]
    \begin{center}
        \includegraphics[width=0.7\linewidth]{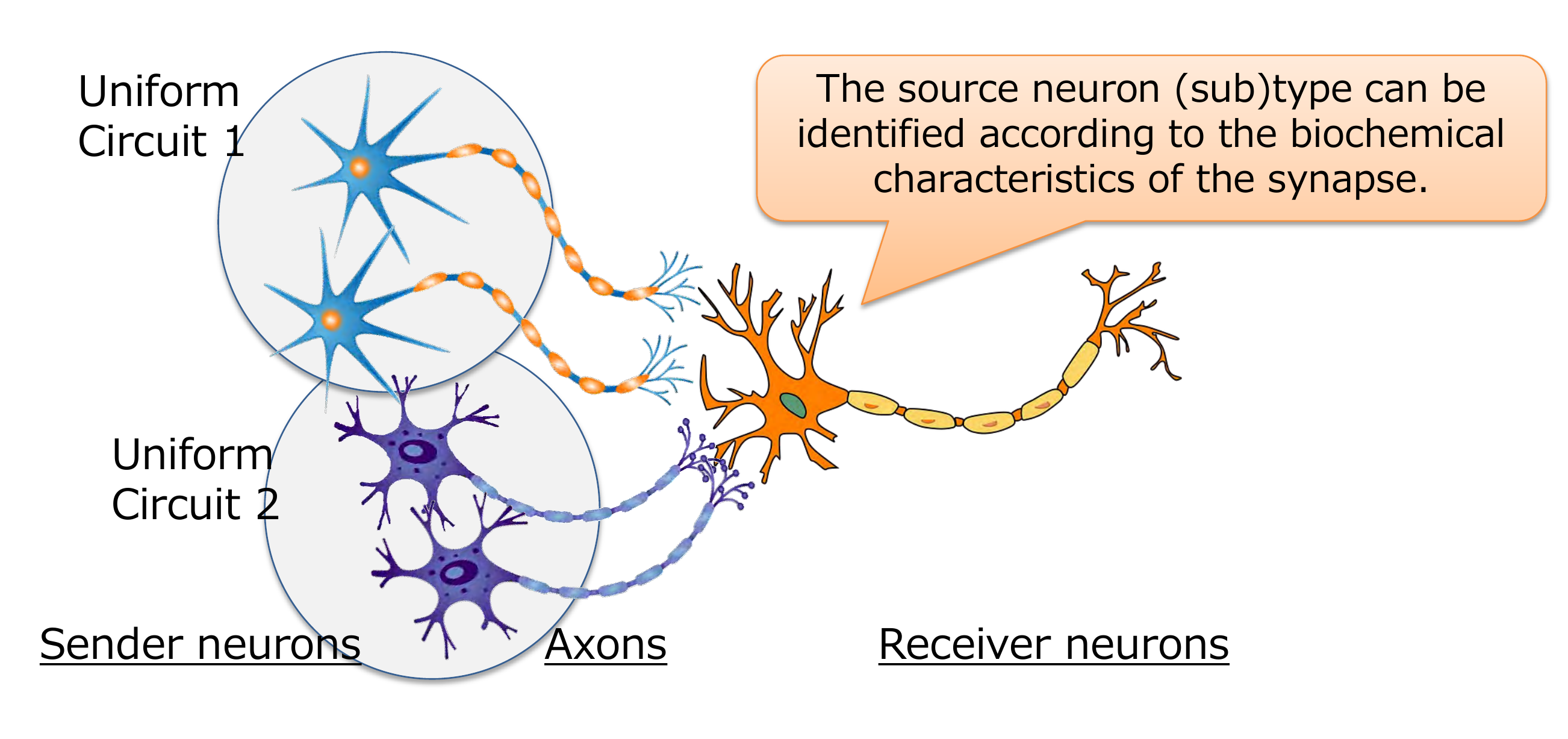}
    \end{center}
    \caption{Uniform circuits.
    Uniform circuits are composed of the same type or subtype of neurons that are spatially localized. Each neuron in the transmitting uniform circuit that extends axons to the receiving neuron will be recognized as physiologically identical on the receiving side.
    }
    \label{fig:4}
\end{figure}

In component diagrams that attempt to associate neural circuits, the smallest unit to be described is the argument specified in the interface of each component (see the right side of Fig. \ref{fig:6}). As an important property of this argument, the transmitting side and the receiving side have a transmitted signal in common. In communication between living neurons, the receiving neuron biochemically identifies its type or subtype by coming into contact with the axon tip of the transmitting neuron (see Fig. \ref{fig:4}). In this regard, a mostly homogenous neuron population belonging to the same brain region is defined as a uniform circuit. Then, because it is a uniform circuit, it can function as an argument in the software.

In well-studied regions of the brain, there are some known examples wherein variables that describe computational models related to cognitive-behavioral function match the uniform circuit.
It is often suggested that this level of anatomical  granularity  has been implicated in the construction of cognitive and behavioral functions across species \citep{Bohland2009-ce, Oh2014-wp}.
 Therefore, in the WBA approach, we decided to essentially refer to neuroscientific findings at the mesoscopic level with the uniform circuit as the lower limit of descriptive granularity.

\subsection { Brain information flow  }
\label{subsec:BIF}

\begin{figure}[tb]
    \begin{center}
        \includegraphics[width=0.95\linewidth]{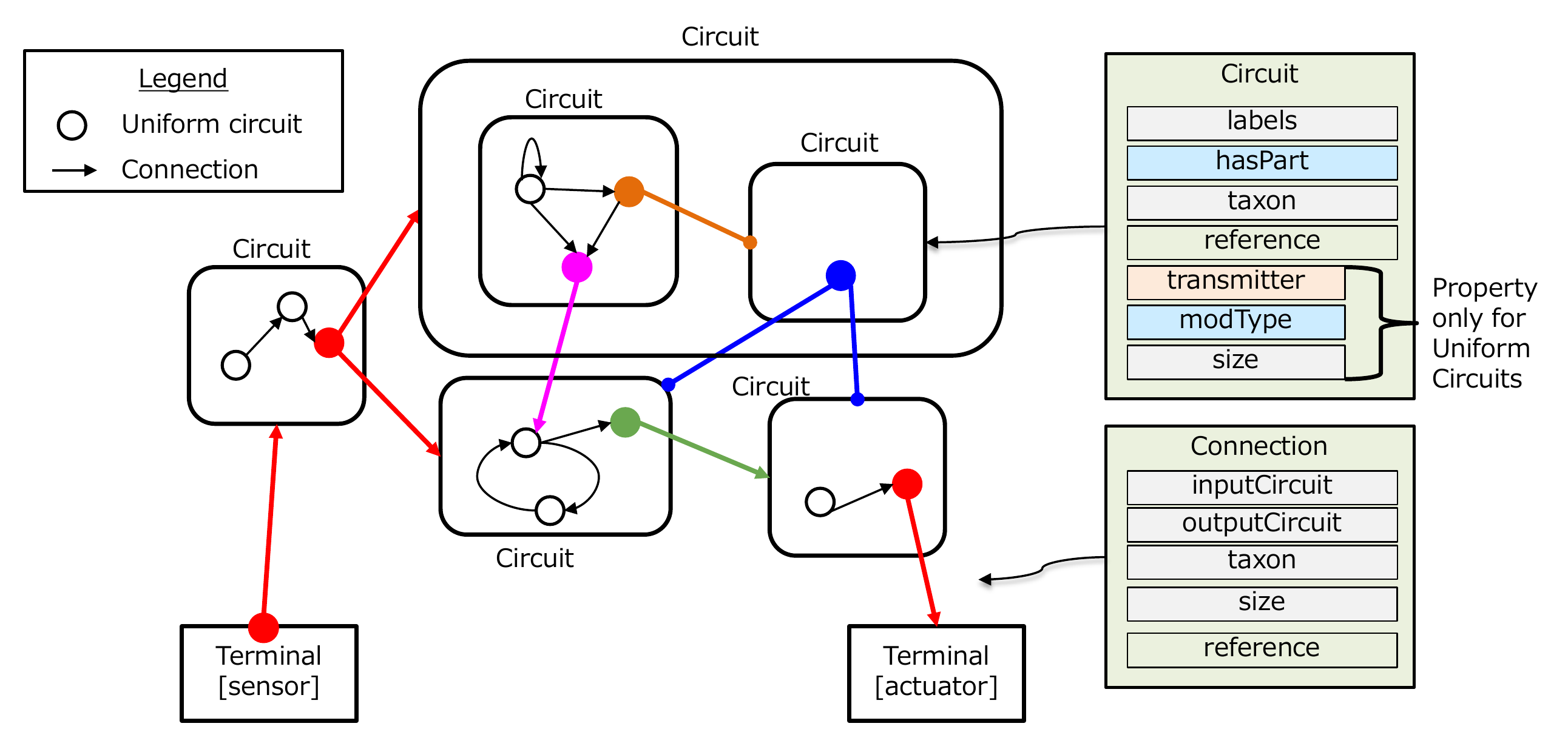}
    \end{center}
    \caption{Brain information flow.
    BIF is the flow of information at the mesoscopic level in anatomical structures. Circuits are arbitrary brain organs or regions and the assemblies to which they are connected. The smallest circuit to be described is, in principle, a uniform circuit. Because a uniform circuit is a collection of neurons of the same type, its properties, such as transmitters, are specified. A connection can be characterized by a particular uniform circuit that is the origin of the connection output from a circuit.
    }
    \label{fig:5}
\end{figure}

The BIF describes the anatomical structure of the whole brain at the mesoscopic level \citep{Arakawa2020-ep} (see Fig. \ref{fig:5}). As such, it is not intended for specific tasks in the environment.  BIF is a graph, and its basic structure consists of a node called a "circuit" and a directed link called a "connection." The smallest unit of the graph is the uniform circuit defined in Subsection \ref{subsec:Mesoscopic-level}, and this unit is also the ideal starting point for a connection. A circuit is also a graph containing multiple uniform circuits. Furthermore, multiple circuits may have overlapping portions.

The circuits and connections are discussed below.
\subsubsection{Circuits (nodes)} 


A circuit is a component that becomes a node in the graph structure of the BIF. A uniform circuit is a special circuit that is the lower limit of the BIF granularity. A circuit can generally be any sub-circuit in the brain. It may indicate areas such as the entire visual cortex or only V1 (i.e., the primary visual cortex), or it may correspond to the neocortex-basal ganglia loop. The attributes possessed by all circuits include labels, partial circuits, animal species, and references. Furthermore, unique attributes of the uniform circuit include neurotransmitters, excitatory and inhibitory modes, and cell count.

\subsubsection{Connections (links) }


Connections correspond to a bundle of axons responsible for signal transduction between circuits in the brain and are represented by a direct link. The main attributes possessed by one connection is the input circuit, which is a uniform circuit that transmits signals, and the output circuit, which sends signals from the connection. Furthermore, animal species, the size corresponding to the number of axons, neurotransmitters, and references can be added to the connection's description if necessary.

\subsection { Hypothetical component diagram }
\label{subsec:HCD}

\begin{figure}[tb]
    \begin{center}
           \includegraphics[width=0.9\linewidth]{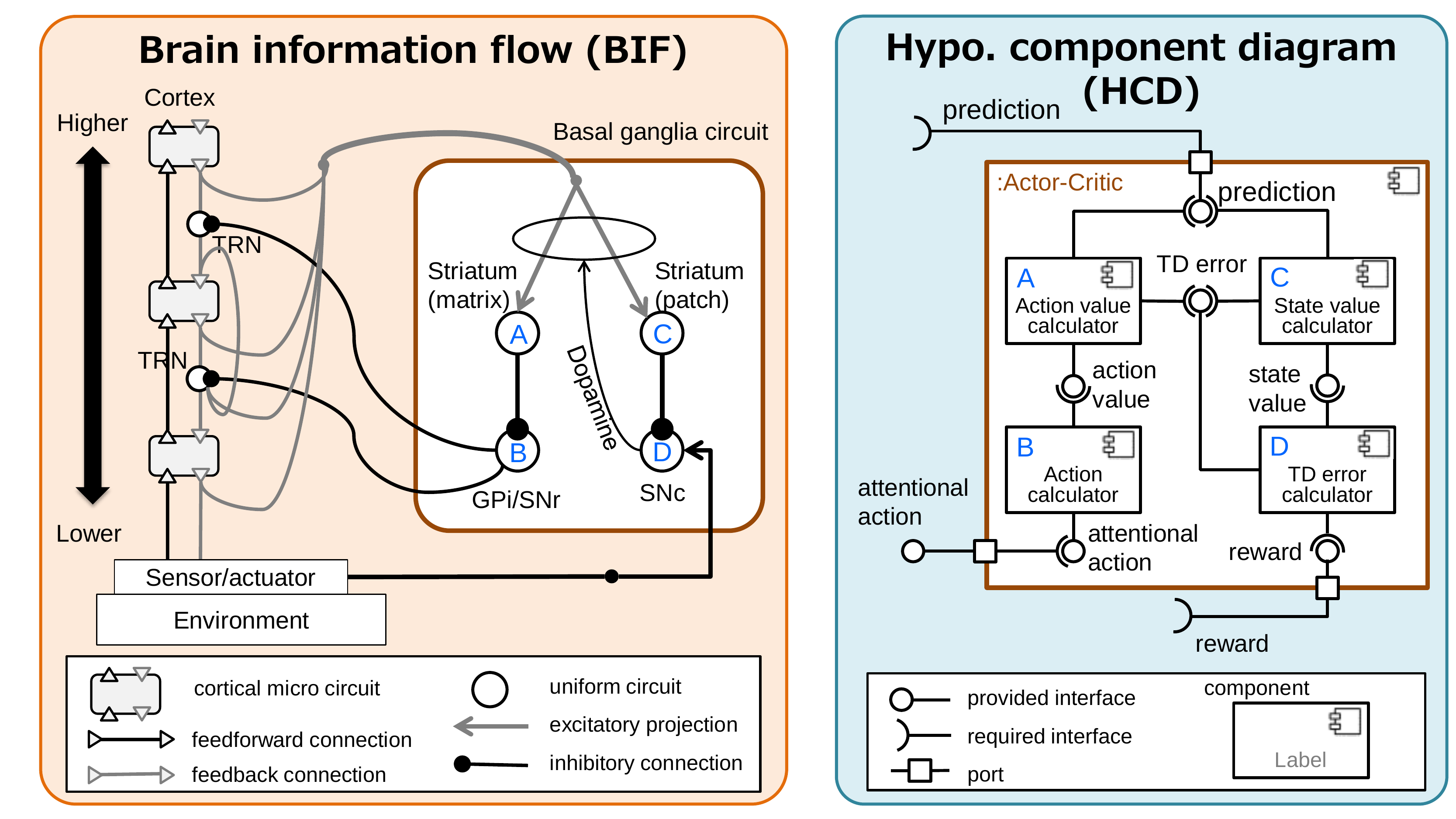}
    \end{center}
    \caption{Example of BRA description
    that associates the BIF for the basal ganglia with the HCD for actor-critic reinforcement learning. The blue letters A, B, C, and D represent uniform circuits in the region of interest (ROI) in the BIF (left panel). The corresponding components in the HCD (right panel) are marked with the same letters, and the HCD components are marked with their functions. The meaning of the signal to be transmitted is indicated with the interface between the components. In the BIF, TRN indicates a thalamic relay neuron. This diagram was adapted from the diagram of \cite{Takahashi2008-eo, Yamakawa2020-oz}.
    }
    \label{fig:6}
\end{figure}

The component diagram of the Unified Modeling Language \citep{Ambler2004-cz} is used to model and explain the structure of any complex object-oriented software. This is a diagram that shows the structural aspects of the operating principle of software as a network using socket labels to indicate the components that are responsible for the computing functions; it also shows the interface of the call relationship between those components (see the right image in Fig. \ref{fig:6}). This diagram is used to visualize, specify, document, and build an executable system by forward or reverse engineering.


The HCD that makes up the BRA is a type of component diagram. This diagram assigns components that match the function of the brain region of interest (ROI) with an anatomical structure at a mesoscopic level; however, this diagram is hypothetical in the sense that there is no guarantee that it is consistent with the truth of the brain function. The assignment is performed using the SCID method, which will be discussed later. Each component that makes up an HCD is a module that encapsulates a set of related functions (or data) \footnote{The term "component" is also used in software engineering to mean software packages, web services, web resources, and similar entities; however, in this paper, we use it to refer to a module that encapsulates a set of related functions (or data).} and corresponds to the behavior and structure of specific organs and regions of the brain.



As a typical example of a BRA, the association between BIF and the HCD is shown in Fig. \ref{fig:6}, which shows the well-known example \citep{Takahashi2008-eo} of the basal ganglia's actor-critic reinforcement learning function. In the left diagram showing the basal ganglia loop, the basal ganglia circuit is the ROI. The corresponding HCD that decomposes the actor-critic reinforcement learning function is shown on the right side. The uniform circuit named striatum (matrix) and indicated by the letter A in the BIF diagram corresponds to the action value calculator component indicated by the letter A in the HCD. The uniform circuits in BIF similarly correspond to the components in the HCD, as indicated by other letters (B, C, and D). The following is an example of mapping the links between the two diagrams. The signal path (labeled dopamine) output from the SNc, indicated by the letter D in the BIF, maps to the signal path (labeled TD error) output from the TD error calculator component, indicated by the letter D in the HCD. Note that in this example, the circuits indicated by A, B, C, and D in the BIF are all uniform circuits. Therefore, the label names given to the respective components in the HCD correspond to the label names of the arguments they provide.


In this way, the availability of an HCD, which shows the structural aspect of operating principles, increases the likelihood that even developers without profound expertise in neuroscience would be capable of implementing software closer to the truth of the brain. Network machine learning systems that are often used in current AI research (e.g., artificial neural networks and Bayesian networks) are made compatible with development according to component diagrams.

\subsection { Prototype of BRA database }
\label{subsec:BRA-DB}


The brain is a fairly closely linked system. Therefore, the accumulation of standard neuroscientific findings related to cognitive behavior will not only optimize the development of brain-inspired software but also help to comprehensively grasp mesoscopic findings in the whole brain.


In this regard, the WBA approach examines databases to improve reusability by integratively handling the constructed BRA. In the present study, a prototype of the BIF database was made using Semantic MediaWiki.


The data flow was as follows. First, one of the authors with expertise in neuroscience reviewed academic papers and collected relevant data in a spreadsheet. Next, the data were registered in a database using a conversion tool. Thereafter, when the developers implemented the brain-inspired software, a prototype of a tool was made, which could not only directly browse the data but also visualize BRA data as a graph in the ROIs.

Such activity also can be positioned as part of the field of neuroinformatics \citep{Amari2002-rp,Pradeep2013-gb}, which develops data and knowledge  bases for neuroscience.
Currently, that field is vigorously registering experimental data on anatomical structures \citep{Kuan2015-tl}  and physiological phenomena \citep{Poldrack2017-ni}.
However, there has been no progress in accumulating data for designing cognitive and behavioral functions, such as a BRA.

\subsection { BRA-driven development}
\label{subsec:BRA-driven-development}

\begin{figure}[tb]
    \begin{center}
        \includegraphics[width=0.99\linewidth]{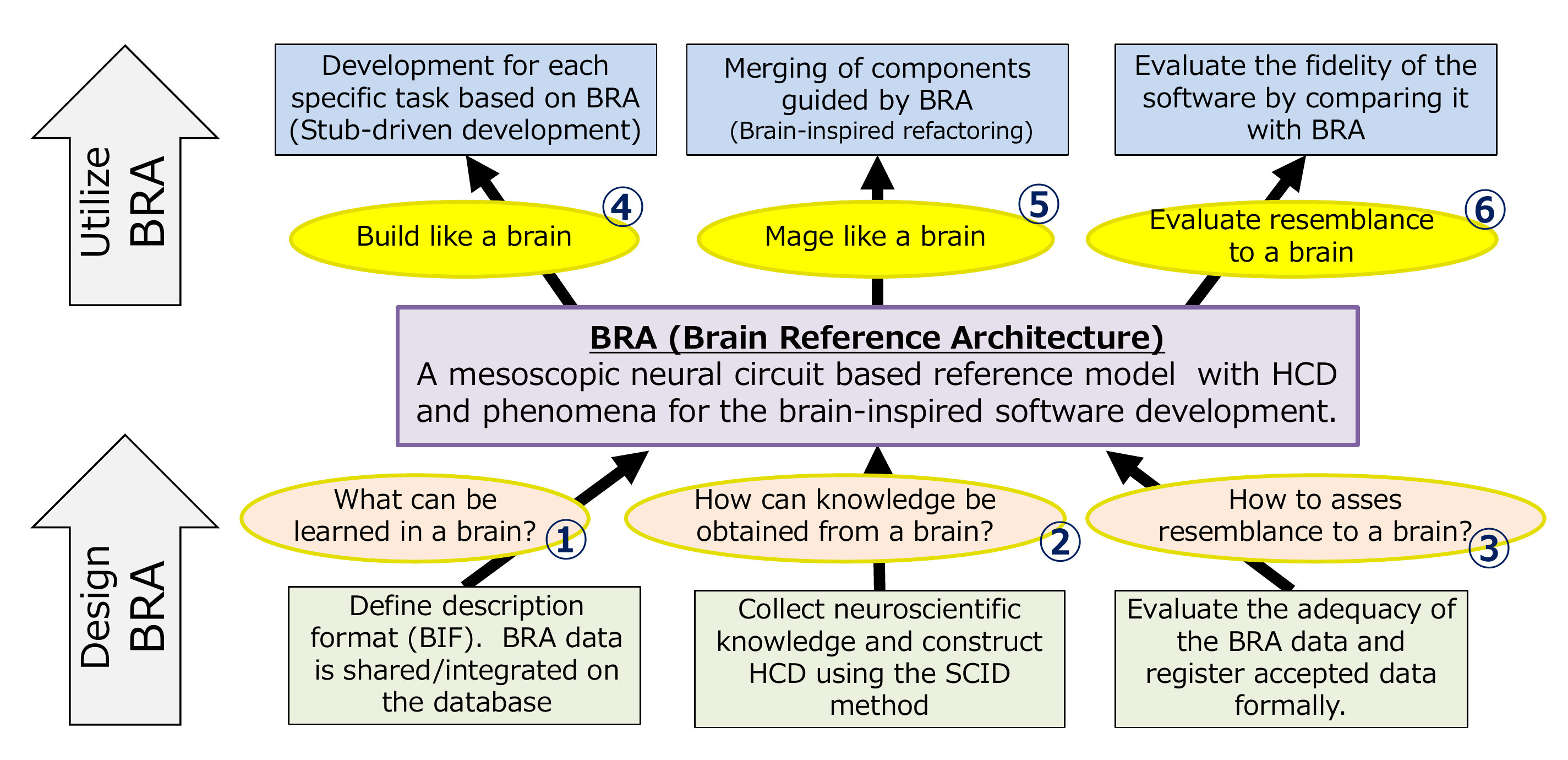}
    \end{center}
    \caption{Activities of BRA-driven development.
    The major activities of BRA-driven development include three types of activities for designing BRA and three types of activities for using BRA.
    }
    \label{fig:7}
\end{figure}

BRA-driven development is a developmental approach that builds brain-inspired AGI through the following processes using a standardized BRA (see Fig. \ref{fig:7}).

\begin{itemize}
  \item BRA design: Designing BIF by collecting and organizing neuroscientific findings. Furthermore, creating an HCD using the SCID method (discussed later).

  \item BRA use: Developing brain-inspired software by referring to the HCD in the BRA.
\end{itemize}

By doing so, the developer can develop brain-inspired software guided by HCDs in the BRA and can compensate for the lack of people who have expertise in both neuroscience and software engineering.


Here, we will give an overview of the three activities involved in BRA design.
The first activity is laying the foundation for accumulating BRAs useful for brain-inspired software. Specifically, this involves deciding on the description format (BIF, Subsection \ref{subsec:BIF}) and examining databases for integrated sharing of BRAs (Subsection \ref{subsec:BRA-DB}). (① What can be learned in a brain?)
The next activity is acquiring and formulating the knowledge necessary for the BRA.  Specifically, knowledge of the anatomic structures and psychological phenomena in some region of the brain is collected and amassed in BIF form. Those BIF data are used to construct HCDs through the SCID method. (② How can knowledge be obtained from the brain?)
The third activity is evaluating the adequacy of BRA data. In this activity, review criteria are determined, and one judges whether the created BRA data satisfy the necessary requirements as a reference model of brain-inspired software. (③ How to assess the resemblance to a brain.)

Furthermore, we will give an overview of three activities associated with BRA utilization.
First, in the development of brain-inspired software, the HCD associated with a specific task in the BRA is implemented as a requirement. (④ Build the software like a brain.)
Next, in the future, we plan to carry out merge development wherein components in separately developed programs are associated with each other based on BRA and integrated. (⑤ Merge disparate functions as the brain does.)
Furthermore, to estimate how well the implemented software represents the brain, the fidelity (biological plausibility) is evaluated by comparing the BRA and the program. (⑥ Evaluate the resemblance to a brain.)


As previously mentioned, to complete the brain-inspired AGI, it is necessary to integrate multiple computational mechanisms that correspond to the same brain regions and were created according to, for example, diversity of tasks, in each BRA-based development project. Therefore, we believe that the entire development based on the WBA approach in the near future will proceed in parallel or iteratively with BRA-based developments and merge developments. In that case, the fidelity evaluation of the software will prevent development results from veering away from brain architecture.

\section {Designing BRA}
\label {sec:Designing-BRA}


In this section, we discuss the BRA design. Among the three activities related to the BRA design, the first part, "① What can be learned in a brain?" was already discussed in Section 2. As for "② How can knowledge be obtained from a brain?" we will explain the collection of anatomical findings in the neuroscientific field and the SCID method for HCD construction. Furthermore, we will also touch on the evaluation of adequacy in the context of "③ How to assess resemblance to a brain."

\subsection { Neuroscientific findings available for BIF creation }
\label{subsec:Neuroscientific-findings}


We now discuss the process of acquisition of information related to the anatomical structure required to describe the BIF. The main requirement is information for building directed graphs with circuits as nodes. Therefore, it is ideal to acquire information on uniform circuits of the whole brain and the connections between these circuits. With the current state of neuroscience, we are still far from acquiring ideal information. In this regard, if necessary, circuits larger than the uniform circuit are defined, and a BIF graph is constructed between these circuits.


Information to be acquired for each uniform circuit includes the brain region labels, animal species, neurotransmitters, excitatory and inhibitory modes, cell count, and information source (references). The information to be acquired regarding the connection includes the input circuit, output circuit, animal species, size (number of axons), neurotransmitters, and sources (references). For the neocortex, the orientation of the hierarchy between territories (including feedforward/feedback) is required.


Furthermore, the data described in the BIF is used to implement software using an artificial neural network. Thus, it is ideal to know the number of neurons in a circuit and the approximate connection sizes (number of axons).


A BRA used to build human-like intelligence should, of course, be based on the structure of the human brain. However, streamlining the construction of the BIF may be possible by referring to the findings in other animals, particularly rodents. For this reason, in reality, the BIF mainly uses human data for the neocortex, which is unique to humans; however, for other brain regions, there are several references based on non-human primates and rodents \citep{Negishi2019-pt}. Therefore, although similar to humans as a whole, the BIF seems to contain chimeric data that combines mesoscopic-level anatomical findings from multiple mammals. 

\subsubsection{Information sources}


The main source of information for building a BIF is data on anatomical structures (such as connectomes) and related literature.
\subsubsection {Region labels}


In principle, the Allen Developing Human Brain Atlas ontology, which is one of the Allen Brain Reference Atlases (https://atlas.brain-map.org/), is used as a circuit label, and if necessary, a label with a level roughly corresponding to the granularity of the uniform circuit is added.
\subsubsection { Number of neurons}


The number of neurons for each region in the mouse brain is stored in the Blue Brain Cell Atlas \citep{Ero2018-rt}.  These regions are defined by the Allen Mouse Brain Reference Atlas \citep{Kuan2015-tl}. However, to date, no comprehensive data on the number of neurons in humans are available.

\subsubsection {Connection}


As for the connections between circuits, we would like to gather information on the presence of the connection, its direction, and the approximate number of projection axons for all combinations of areas.


Although that is not necessarily an exhaustive brain region, we can estimate the projection ratio from one particular area to another using the Allen Mouse Brain Connectivity Atlas \citep {Oh2014-wp}.  As mentioned above, because the number of neurons in an area can be obtained for mice, the number of axons to be projected can be estimated by multiplying the projection ratio by the number of neurons.


In humans, the Multilevel Human Brain Atlas by EBRAIN
 (https://ebrains.eu/service/human-brain-atlas/ .  Accessed:  2021-2-26.) can be used to obtain data, including hierarchical relationships (feedforward/feedback), for the entire neocortex.

\subsubsection { Neurotransmitters}


The data on the distribution of neurotransmitters throughout the brain is currently available via Drosophilidae studies \citep{Meissner2019-wx}, but there appears to be no data for mammals. However, in the regions involved in higher intelligence processing, such as the neocortex, thalamus, basal ganglia, hippocampus, and cerebellum, similar anatomical structures appear frequently in each brain region. Moreover, because the neurotransmitters in these sites have been studied in detail, the lack of data does not pose a major problem in the construction of the BIF. Nevertheless, data for subcortical brain regions may be needed.

\subsection { Structure-constrained Interface Decomposition method }
\label{subsec:SCID-method }


The SCID method involves consistently decomposing the computational functions of a specific brain region into mesoscopic-level anatomy to obtain the HCD required for the development of brain-inspired software. In software development, designing through a process of decomposing higher-level functions is common; however, the SCID method also considers consistency with the anatomical structure of the brain.
	

Furthermore, decomposing the functions of the natural brain as if it were an artifact might not yield desired results. However, because the brain is an organ that has undergone evolutionary selection, its physical mechanism often serves an intended purpose. For example, when computational neuroscience derives "algorithms and expressions" for brain function, this action is premised on a clear purposefulness.

\subsubsection{Process of SCID method}

\begin{figure}[tb]
    \begin{center}
        \includegraphics[width=0.99\linewidth]{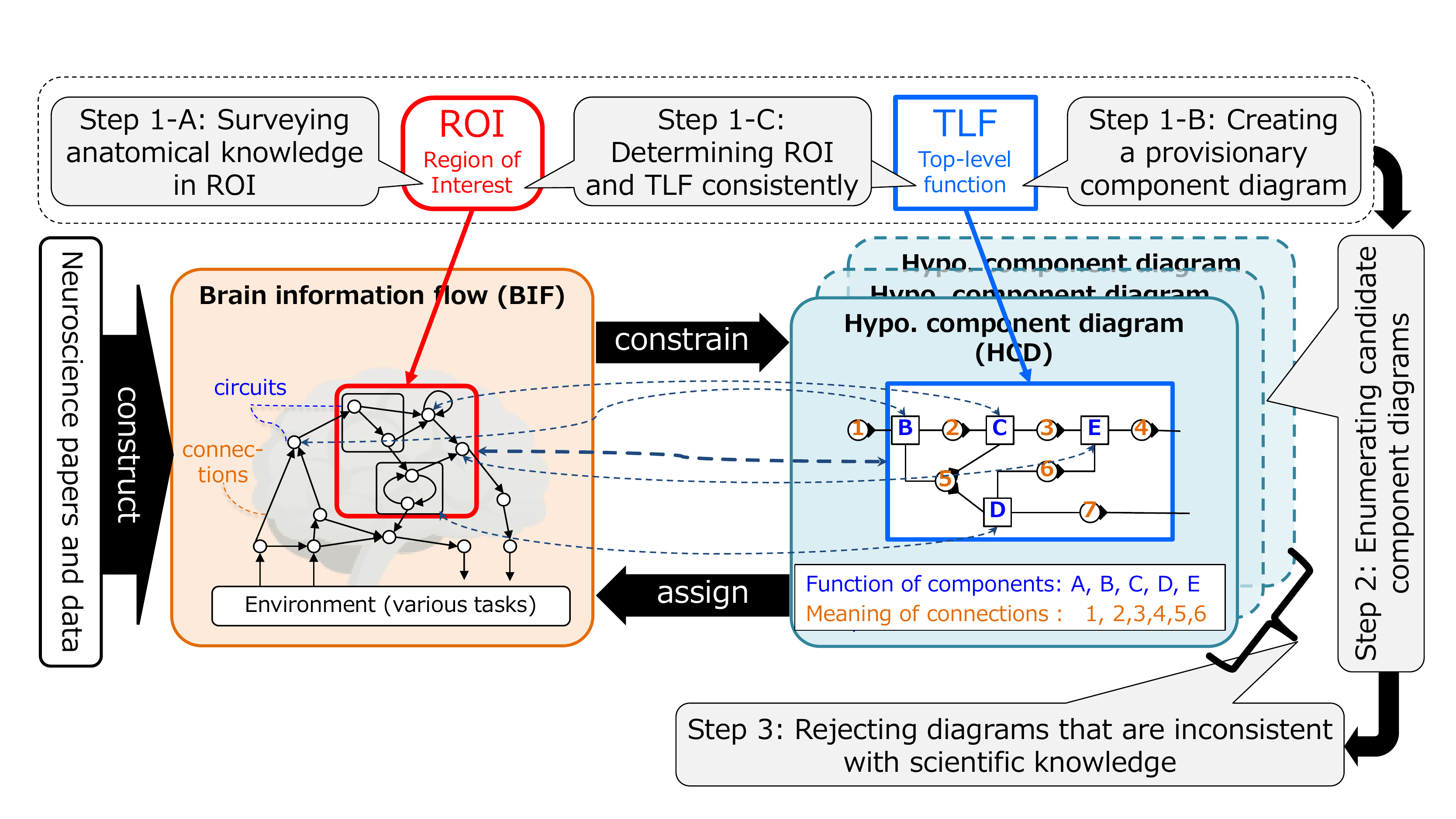}
    \end{center}
    \caption{Procedure  of SCID method.
    The SCID method is a technique for obtaining the HCDs required for software development. The method uses a three-step process to decompose the top-level functions (TLFs) of a specific brain region (ROI) into components under mesoscopic-level anatomical constraints.   In Step 1, three exploratory tasks (1-A, 1-B, and 1-C) are performed in parallel as appropriate. After that, we proceed to Steps 2 and 3.
    }
    \label{fig:8}
\end{figure}

In the SCID method, an HCD consistent with the anatomical structure in the ROI is obtained by performing the following three-step process (see Fig. \ref{fig:8}).

In step 1, the findings of various studies related to cognitive behaviors of humans and animals are used to establish the premise that the SCID method is applicable. Specifically, the three processes are performed in parallel. While investigating the anatomical structure around the ROI and registering it as a BIF (1-A), the existence of a component diagram that realizes the ROI input/output (1-B) is confirmed. A valid brain ROI and the top-level function (TFL) that it performs are determined (1-C).


In step 2, while considering the association between circuits and connections in the ROI of the BIF, the TLF is decomposed into detailed functions in as many conceivable patterns as possible. This step enumerates candidate HCDs.


In step 3, HCDs that are logically inconsistent according to scientific findings in various fields, such as neuroscience, cognitive psychology, evolution, and biological development, are rejected.  Then, the function of components and meaning of connections of the remaining HCDs can be assigned to the BIF.

\subsubsection{Advantages of SCID method}


In neuroscience, the traditional way to experimentally identify the function of a neural circuit of interest is as follows. This is a method of finding neural activity that has an intelligible correlation with some external stimulus, and giving a functional interpretation to it based on the nature of that stimulus. However, this is only possible if there are brain regions close to the sensor/actuator or neural activity that has a clear correlation with the behavior, such as reward/place cells. In general, it is not easy to obtain interpretable correlations from neural activities that are mixed with various types of external and temporal information in most parts of the brain's neural circuitry. Thus, there is a weakness in that the range of neural circuits with functions that can be identified by this method tends to be limited (see Table 1).


The SCID method can be applied to a quite wide area of the brain. This is because anatomical structure information at the mesoscopic level, which is key to the SCID method, can be obtained from almost the entire brain, including that of rodents (see Subsection \ref{subsec:Neuroscientific-findings}).


Another advantage of the SCID method is that an HCD is easy to use directly in software development because it is obtained through a process based on the design theory of software development. In contrast, when neural activity phenomena correlated with external information are used as a reference for software development, they need to be reinterpreted as a requirement. In other words, the functions obtained through the traditional phenomenon-based approach \citep{Yamakawa2017-oc} often are indirect information and need preparation for software development.

\begin{table}[htb]
  \begin{center}
    \caption{Advantages of the SCID method}
    \begin{tabular}{p{0.15\linewidth} p{0.4\linewidth} p{0.4\linewidth}} \hline
      Method & SCID method & Conventional method \\ \hline 
      Key clues &	Structure and top-level function
      (also physiological phenomena) & Neural phenomena correlating 
      with the environment (e.g., reward and place cells) \\ \hline
      Coverage in the brain	& Almost all of the brain (to the extent that mesoscopic structures are known) & Limited to areas where physiological clues exist. \\ \hline
      Features & Functional descriptions that are easy to use for development & Phenomenal interpretations are indirect and software specific. \\ \hline 
    \end{tabular}
  \end{center}
\end{table}


The first HCD developed using the SCID method identified the site responsible for path integration in the entorhinal cortex \citep{Fukawa2020-hl}. Following this, it was also used to identify the meaning of signals between neocortical regions \citep{Yamakawa2020-xa}. At present, there are several ongoing studies on the application of the SCID method to study brain regions, such as the brain stem, which is responsible for eye movements \citep{Tawatsuji2020-ym}; the claustrum; and functions, such as imagination.
\subsection { Evaluation of adequacy (biological plausibility) }
\label{subsec:Adequacy}

\subsubsection{Need to evaluate biological plausibility}


When developing brain-inspired software, it is necessary to evaluate biological plausibility---in other words, to estimate how close the implemented brain-inspired software is to the reality of the brain captured by current neuroscience findings.

\begin{figure}[tb]
    \begin{center}
        \includegraphics[width=0.8\linewidth]{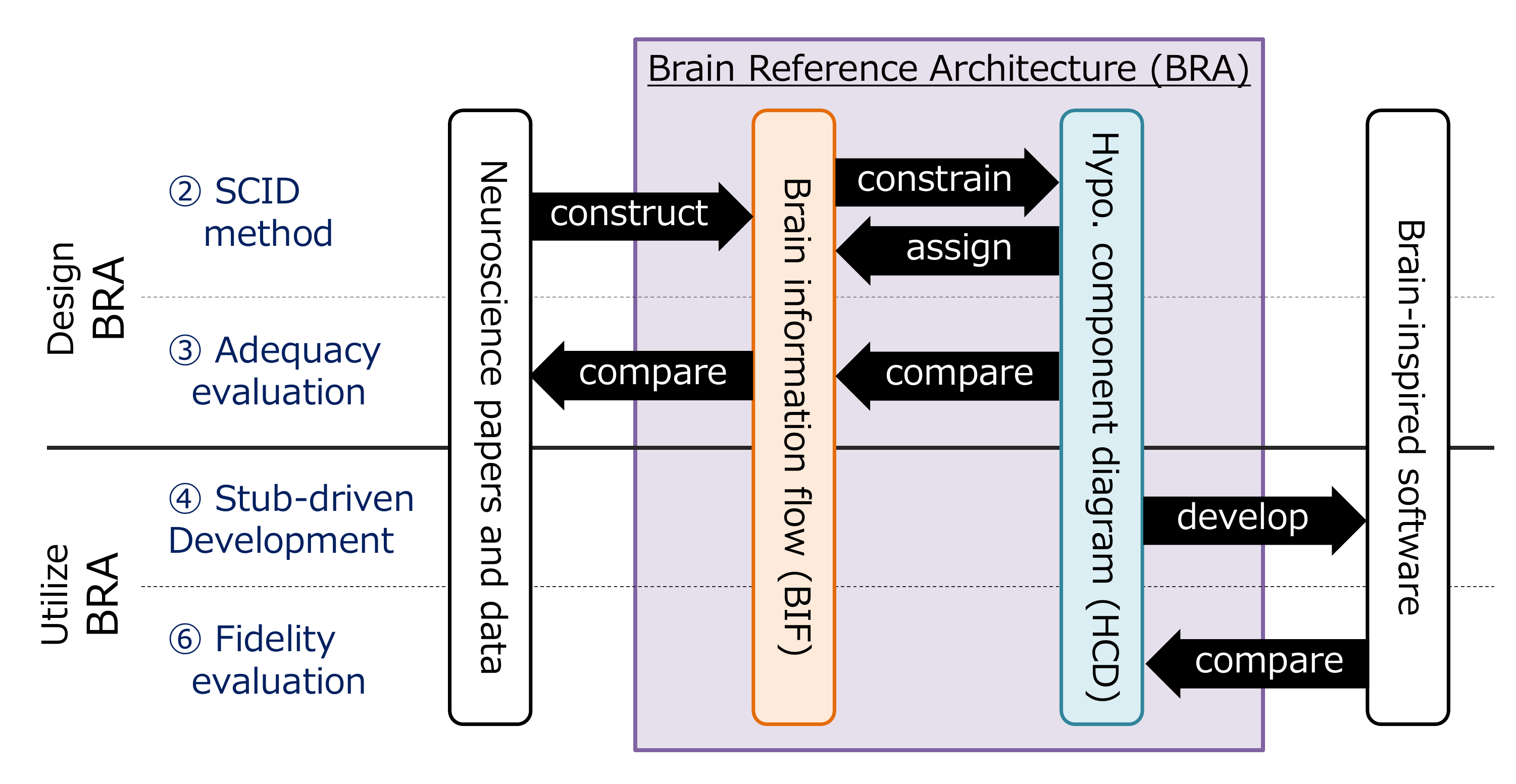}
    \end{center}
    \caption{Creation and evaluation in BRA-driven development.
    In BRA-driven development, the biological plausibility is evaluated in the direction opposite to that of creation. Contrary to the SCID method of designing BRAs, the adequacy of BRAs to neuroscientific findings is assessed. The fidelity of the software to the BRA will be assessed, contrary to the development of the software. Although not shown in the diagram, in merge development, an HCD is used for both development referencing and fidelity evaluation.
    }
    \label{fig:9}
\end{figure}

The evaluation of biological plausibility in BRA-driven development involves the following two methods (see Fig. \ref{fig:9}). The first method is the evaluation of adequacy, which estimates the consistency between existing neuroscientific findings and BRA. The second method involves the evaluation of fidelity, which estimates the reproducibility of BRA in brain-inspired software.

\subsubsection{Need for certified registration}


The created BRA is used as a functional requirement to be referred to when implementing software and as a subject for comparison when evaluating biological plausibility (fidelity). However, most BRA users have little knowledge of neuroscience, and therefore cannot judge whether the BRAs created are trustworthy. To ensure the adequacy of BRA data, there needs to be a workflow that inspects and certifies the data before they are registered.


Often in neuroscience, there are parallel hypotheses that are contradictory to each other but cannot be ruled out. From the viewpoint of brain-inspired software development, it is not possible to immediately determine which hypothesis is ideal. Therefore, provided that it meets the inspection criteria, the BRA data should be registered even if they contradict other data.

.

\subsubsection{Evaluation of adequacy and inspection criteria}


The evaluation of adequacy is further divided into two parts as shown in Fig. \ref{fig:9}.
.

{\bf 1) Adequacy evaluation of BIF }


This method evaluates the consistency of the anatomical structures and neural activity described in the BIF with those described in neuroscientific papers and data.


Two main inspection criteria are used to certify that the BIF description is sufficiently adequate. One is that the description element of the structure or phenomenon described in the data submitted for registration is not already registered in the BRA database (novelty). The other is that the element must be directly or indirectly supported by any current neuroscientific finding (authenticity). As a rule, authenticity of facts is guaranteed by their inclusion in one or more peer-reviewed articles.


{\bf 2) Adequacy evaluation of HCD }

The functionality of and consistency with the BIF are evaluated for the HCD. The evaluation of functionality determines whether the process generated by the behavior of structured components in an HCD constitutes a mechanism of action that can achieve the goals of the ROI.


The evaluation of consistency determines whether the HCD corresponds to the description of the BIF from three aspects:
(1) the dependency structure of the HCD corresponds to the anatomical structure contained in the ROI of the BIF, 
(2) the behavior of the components within the HCD is consistent with the physiological findings described in the BIF, and
(3) the goal achieving process of HCD is consistent with the physiological process described in the BIF.

\section {Development and Evaluation using BRA }
\label {sec:Development&Evaluation }


In this section, we discuss three activities associated with BRA use in BRA-driven development. As shown in Fig. \ref{fig:9}, the development and evaluation performed here is carried out with reference to the HCD in the BRA; as such, the programmer does not need to possess profound knowledge of neuroscience.
\subsection { Stub-driven development}
\label{subsec:Stub-driven-development}


To create brain-inspired software, BRA-driven development implements and connects all components based on the requirements HCD associated with a particular task.


Generally, machine learning devices often behave differently from the architecture imagined at the design phase to some extent. The difficulty of controlling this behavior increases rapidly if the system is composed of several machine learning components. To address this difficulty, the WBA approach uses stub-driven development.


In stub-driven development, during the early stages of development, a system is constructed by combining components that do not have a learning function and are described by rule-based processes.
Following this, by gradually replacing each component with machine learning components, the system is improved so that it approaches the behavior expected in the HCD.


To proceed with development using a BRA, an integrated execution platform is required as a management mechanism for computational resources to run and train multiple components.
Candidates include recent deep neural network platforms, such as TensorFlow, PyTorch, and Keras.
The Brain-inspired Computing Architecture is a platform developed to take into account the asynchronous nature of the brain and other characteristics \citep{Takahashi2015-ko}.
Furthermore, an HCD can be constrained and converted into a probabilistic generative model (PGM), SERKET \citep {nakamura2017serket, Taniguchi2020-yb}, and Pixyz \citep{Pixyz-github}.
Recently, there has been a growing movement called whole-brain PGM, which attempts to construct a PGM corresponding to  the whole brain \citep{Taniguchi2021wb-pgm}. Construction of a PGM of the hippocampal formation has been started \citep{Taniguchi2021hpf-pgm}

\subsection { Evaluation of fidelity (biological plausibility) }
\label{subsec:Fidelity}


The biological plausibility of brain-inspired software is evaluated by comparing it with the BIF and HCD in the BRA data. The estimated degree of consistency between the software and the BRA is called fidelity.


To date, four methods have been explored for evaluating fidelity.

\begin{itemize}
  \item {\bf Structural similarity:} An evaluation of how well the static structure of the software matches the BIF in the BRA.
  
  \item {\bf Functional similarity:} An evaluation of how well the behavior of a particular component implemented during the execution of a specific task matches the behavior (e.g., behavior timing) designed in the HCD in the BRA.
  
  \item {\bf Activity reproducibility:} An evaluation of how well the behavior of a specific component implemented during the execution of a specific task given to a circuit in the BRA reproduces the features of neural activity (e.g., activity timing).
  
  \item {\bf Performance:} An evaluation of the performance and ability of the software as a whole (integration test).
\end{itemize}


Of these evaluation methods, structural similarity and performance are easy to use for evaluation of the entire software. However, functional similarity and activity reproducibility are useful not only for unit tests for each component but also for merge development, which will be discussed later. Furthermore, it is possible to consider an evaluation method wherein dysfunction states are induced by intentionally destroying/ablating a part of the software and comparing it with brain function under conditions such as mental illness or brain injury.

\subsection {Merge development}
\label {subsec:Merge-evelopment }


A particular circuit on the BIF is associated with a component that is included in various HCDs.
As already mentioned, the HCD is a structure of functions broken down into components to realize TLFs, including tasks.
Therefore, even if a component is implemented to realize the same circuit, the function it performs may be different depending on the HCD to which it refers.


However, for a piece of software to show its true value as an AGI, it needs to be able to apply knowledge to different tasks. To that end, if there are components that correspond to the same brain region in separately developed programs, merge development is performed by associating and integrating these components. The idea of promoting the integration of components by using constraints of the brain is called brain-inspired refactoring.
\subsubsection{Concept of merge development}


Herein, we explain that merge development is possible by imposing the BIF as a development constraint.

\begin{figure}[tb]
    \begin{center}
        \includegraphics[width=0.9\linewidth]{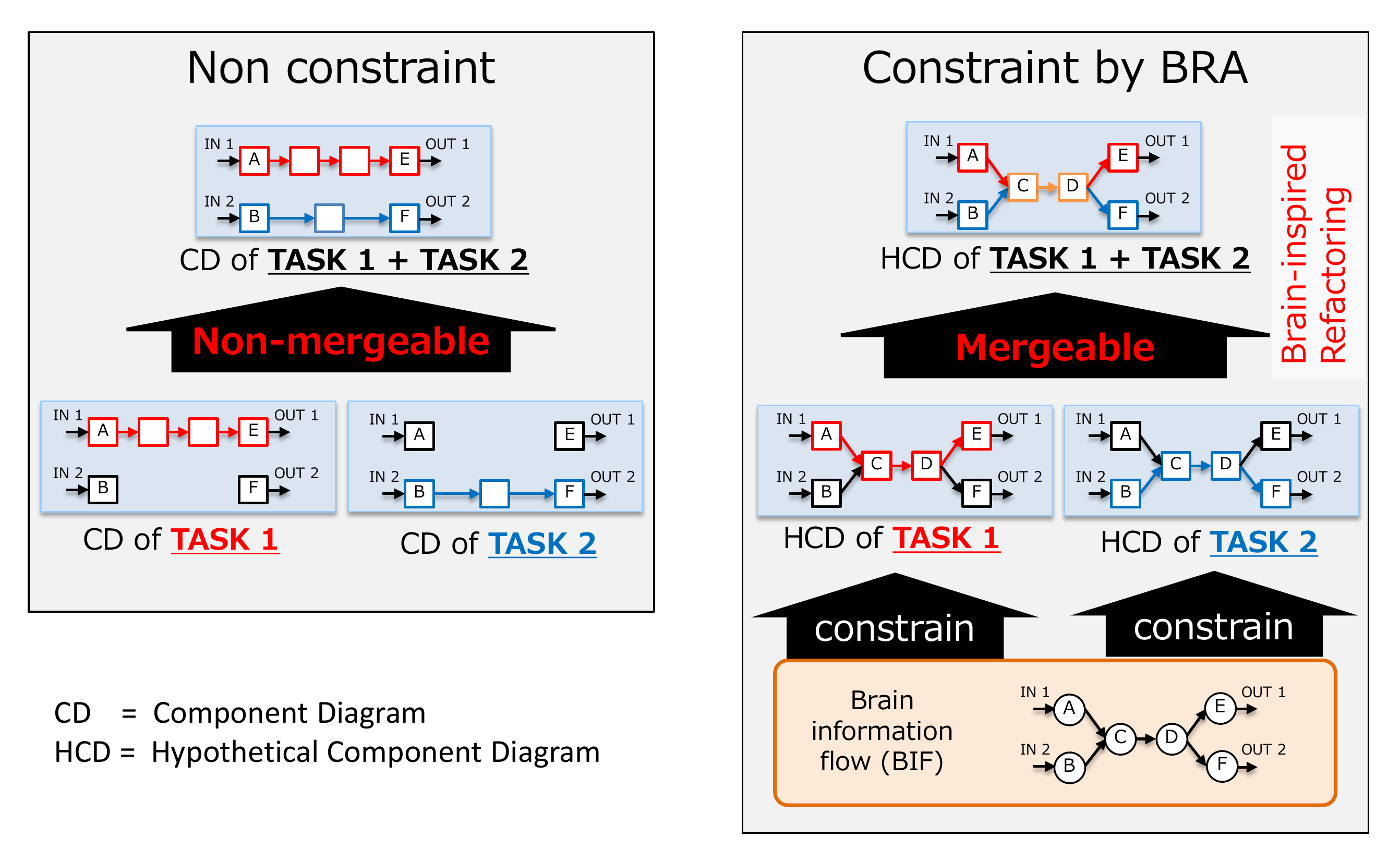}
    \end{center}
    \caption{Components to be merged are identified by BRA constraints.
    This figure compares software development unconstrained by BRA (left panel) and software development constrained by BRA (right panel). Task 1 produces output 1 from input 1, and Task 2 produces output 2 from input 2. Comparing the component diagrams dealing with these two tasks, it is not easy to find common points simply by developing them separately, even if there is a possibility of overlap. However, if they are developed with reference to the BRA, the BIF will constrain the HCD that the software follows. Then, components C and D, which are responsible for common processing in the two tasks, can be identified as components to be merged.
    }
    \label{fig:10}
\end{figure}

As shown in the left side of Fig. \ref{fig:10}, when developing two tasks, the development is done independently if a BRA is not used as a reference. In task 1, input 1 is given to component A, and after processing, output 1 is obtained from component E. Similarly, in task 2, input 2 is given to component B, and after processing, output 2 is obtained from component F. The two implementations created this way are completely different in essence and cannot be merged.

Next, as shown in the right side of Fig. \ref{fig:10}, the case where two tasks are developed with BRA as the constraints are discussed. In this case, components C and D, which are responsible for intermediate processing, are associated with the same circuit on the BIF, respectively.
To merge components C and D contained in different implementations, two approaches are typically envisioned. The first is to compare the fidelity ratings of the two corresponding components and select the higher-evaluated implementation. The second is to redesign and implement an integrated algorithm that combines the advantages of both.

\section {Discussion}
\label {sec:Discussions} 

\subsection{Structure and behavior} 


In BRA-driven development, we decided to use HCD as the main source of design information because it expresses the structural aspects of functions. Moreover, the accumulation of findings has progressed because the mesoscopic-level structure in the brain, which is the basis of an HCD, embodies invariant information that is not task-dependent. However, not only the structural aspect but also the behavioral aspect is important in software design \citep{Ambler2004-cz}. Furthermore, the intelligence finally evaluated in AGI relates to behavior. Therefore, an important task in the future is to develop behavior-related evaluations that are similar to functional similarity and activity reproducibility used for fidelity evaluation.

\subsection{Roadmap for reaching AGI}


As mentioned above, BRA-driven development consists of BRA design comprising the BIF and an HCD and development using them.
Given the characteristics of this development methodology, the following four milestones need to be achieved on the way in the roadmap leading to the realization of a brain-inspired AGI.
The first milestone is the construction of a BIF for almost the whole brain. The second milestone is the construction of an HCD that covers almost the whole brain, and the third milestone is the construction of integrated software that covers almost the whole brain. 
Finally, the fourth milestone is to achieve a situation where AGI solutions can be explored in a design space where constraints continue to be updated as neuroscience progresses, primarily by running and testing integrated software in a virtual environment.
However, something may still be lacking even when all the computational mechanisms that make up the brain seem to work together. Regardless, we expect that the lacking technical elements will become apparent once we arrive at that stage.

\subsection{Applications of AI systems based on BRA }


AI systems developed based on BRA can be expected to almost exactly replicate human cognitive and behavioral capabilities.
Therefore, it has several practical applications, including the following: Its greatest use is that it allows us to construct an AI that has familiarity with humans when communicating with them. Furthermore, it can also be computationally applied to research fields that deal with mental illness and cognitive impairment. Conversely, findings regarding human cognitive impairment may be used for the problematic behavior seen in brain-inspired AI. Moreover, we believe this can also be used as a computational model that will serve as a device for mind uploading.

\section {Conclusions}
\label {sec:Conclusions}



This article introduces the current WBA approach and focuses on BRA-driven development to accelerate the development of brain-inspired AGI.
The BRA includes standardized data that reflect the brain architecture for the purpose of limiting the huge design space required for a human-level AGI that cannot be grasped by one individual's cognitive ability. 
By having BRAs designed by people with expertise in neuroscience, even developers who do not have a deep understanding of the brain can develop brain-inspired software based on BRAs.
First, we discussed that the BRA is a description consisting of a BIF supported by a mesoscopic neural circuit and an HCD consistent with BIF. Next, to compensate for the lack of neuroscientific findings, we introduced the SCID method, which formulates the creation of an HCD consistent with the brain's anatomical structure. Furthermore, even if a BRA is used for development, individual development results tend to diverge depending on the diversity of target tasks. To address this problem, it is planed to introduce merge development, which integrates and brings AGI closer to how the brain works. Moreover, we discussed the evaluation of biological plausibility using BRA to prevent the developed software from veering away from the brain.


The main contribution of this paper on BRA-driven development, having the following features, is the establishment of a methodology for accumulating data on brain constraints in a form that can be used for software development.

\begin{enumerate}
    \item Separation of design information: BRA data can be used in a variety of development projects because they are described in a standard format for software development that does not depend on any particular development environment.
    \item Standardization of description granularity: As a rule, describing BRA data at a coarser granularity than the mesoscopic level reduces the possibility that development will get caught up in details that are unnecessary for the realization of the target cognitive behavioral level.
    \item BRA design: The method of designing computational functions according to anatomy (SCID method) allows BRAs to be created while compensating for the lack of neuroscientific knowledge in a wide range of brain areas.
    \item Tolerance of diversity: Even BRAs that contain mutually contradictory HCDs can be registered if they have a certain level of validity, thereby reducing the risk of overly narrowing the design space to be considered.
\end{enumerate}

The above features of the BRA will provide a scaffold for large-scale whole-brain software development as the comprehensiveness of its data increases. This will allow the brain architecture to exert a centripetal force as an anchor that can efficiently bring about the convergence and eventual completion of the development of human-like AGI, when currently the development results in this field have a tendency to diverge.

\section*{Declaration of competing interest}
The author declares no known competing financial interests or personal relationship that could appear to influence the work reported in this paper.

\section*{Acknowledgements}
I would like to thank 
Naoya Arakawa, 
Koichi Takahashi, 
Naoyuki Sakai, 
Masahiko Osawa, 
Takashi Omori, 
Shinichi Asakawa, 
Kotaro Mizuta, 
Mei Sasaki, 
Hirokazu Kiyomaru, 
Hitomi Sano, 
and Michihiko Ueno 
 for cooperation in developing the methodology for BRA-driven development.
I would like to thank 
Haruo Mizutani,  
Hiroshi Okamoto, 
Yudai Suzuki, 
Naoyuki Sato, 
Taku Hayami, 
and Hiroto Tamura 
for constructing the basic data for the BIF.
I would like to thank 
Kosuke Miyoshi, 
Kotone Itaya, 
Masayoshi Nakamura,  
Tatsuji Takahashi, 
Masanori Yamada,  
Taro Sunagawa, 
Shion Honda, 
Yutaka Matsuo, 
Yuji Ichisugi,
Satoshi Kurihara,  
and Ryutaro Ichise 
for implementing related software.
I would like to thank  
Ayako Fukawa, 
Takahiro Aizawa, 
Yoshimasa Tawatsuji, 
Akira Taniguchi, 
and Ikuko Eguchi Yairi 
for advancing the SCID method.
I would like to thank 
Haruhiko Bito,
Kenji Doya,
Tadashi Yamazaki,
Michita Imai,
Hiroyuki Okada,
and Nobuo Kawakami
for discussions on research and development. 
This work was supported by the Japan Ministry of Education, Culture, Sports, Science and Technology (KAKENHI Grant Number 17H06315, Grant-in-Aid for Scientific Research on Innovative Areas, Brain information dynamics underlying multi-area interconnectivity and parallel processing) and DWANGO Co., Ltd.


\newpage

\bibliographystyle{bst/model5-names}

\end{document}